\documentclass[runningheads]{llncs}
\usepackage{graphicx}
\usepackage{amsmath,amssymb} 
\usepackage{amssymb}
\newtheorem{mydef}{Definition}
\usepackage{subfig}
\usepackage{url}
\usepackage{color}
\usepackage[width=122mm,left=12mm,paperwidth=146mm,height=193mm,top=12mm,paperheight=217mm]{geometry}
\usepackage{soul}
\usepackage{algorithm}
\usepackage[noend]{algpseudocode}

\makeatletter
\def\BState{\State\hskip-\ALG@thistlm}
\makeatother

\makeatletter
\def\blfootnote{\gdef\@thefnmark{}\@footnotetext}
\makeatother

\makeatletter
\newcommand{\algmargin}{\the\ALG@thistlm}
\makeatother
\newlength{\whilewidth}
\algdef{SE}[parWHILE]{parWhile}{EndparWhile}[1]
{\parbox[t]{\dimexpr\linewidth-\algmargin}{%
		\hangindent\whilewidth\strut\algorithmicwhile\ #1\ \algorithmicdo\strut}}{\algorithmicend\ \algorithmicwhile}%
\algnewcommand{\parState}[1]{\State%
	\parbox[t]{\dimexpr\linewidth-\algmargin}{\strut #1\strut}}

\begin{document}

\pagestyle{headings}
\mainmatter

\title{deepsing: Generating Sentiment-aware Visual Stories using Cross-modal Music Translation} 

\titlerunning{-}

\authorrunning{-}

\author{Nikolaos Passalis$^{1*}$ and Stavros Doropoulos$^{2*}$}
\blfootnote{*Equal contribution. Author order determined through a quantum entanglement experiment. No cats harmed during this experiment.}
\institute{$^1$Aristotle University of Thessaloniki, Greece\\
$^2$DataScouting, Greece\\}

\maketitle

\begin{abstract}
In this paper we propose a deep learning method for performing attributed-based music-to-image translation. The proposed method is applied for synthesizing visual stories according to the sentiment expressed by songs. The generated images aim to induce the same feelings to the viewers, as the original song does, reinforcing the primary aim of music, i.e., communicating feelings. The process of music-to-image translation poses unique challenges, mainly due to the unstable mapping between the different modalities involved in this process. In this paper, we employ a trainable cross-modal translation method to overcome this limitation, leading to the first, to the best of our knowledge, deep learning method for generating sentiment-aware visual stories. Various aspects of the proposed method are extensively evaluated and discussed using different songs.

\end{abstract}

\section{Introduction}

Music is closely tied to human evolution, with various musical instruments, such as flutes, dating back at least 40,000 years~\cite{wade1979music}, while music itself can be traced back even before the Paleolithic era~\cite{morley2003evolutionary}. Compared written and spoken language, music cannot (and does not aim to) accurately and concisely transfer semantic and quantitative information. In that sense, it seems like it has no functional role in our life, since it cannot be used to communicate for practical matters. Despite this, it excels at performing another function: \textit{conveying emotions}. In fact, psychologists and neuroscientists suggest that music had a critical role in developing  human societies, since it significantly assisted the process of socialization~\cite{cross2010evolutionary}. Indeed, music consists an indispensable part of our life in modern societies, with the typical listener spending more than 30 hours per week listening to music~\cite{report}.

\begin{figure}[t]
	\begin{center}
	\includegraphics[trim={0cm 0cm 0cm 0cm},clip, width=0.95\linewidth]{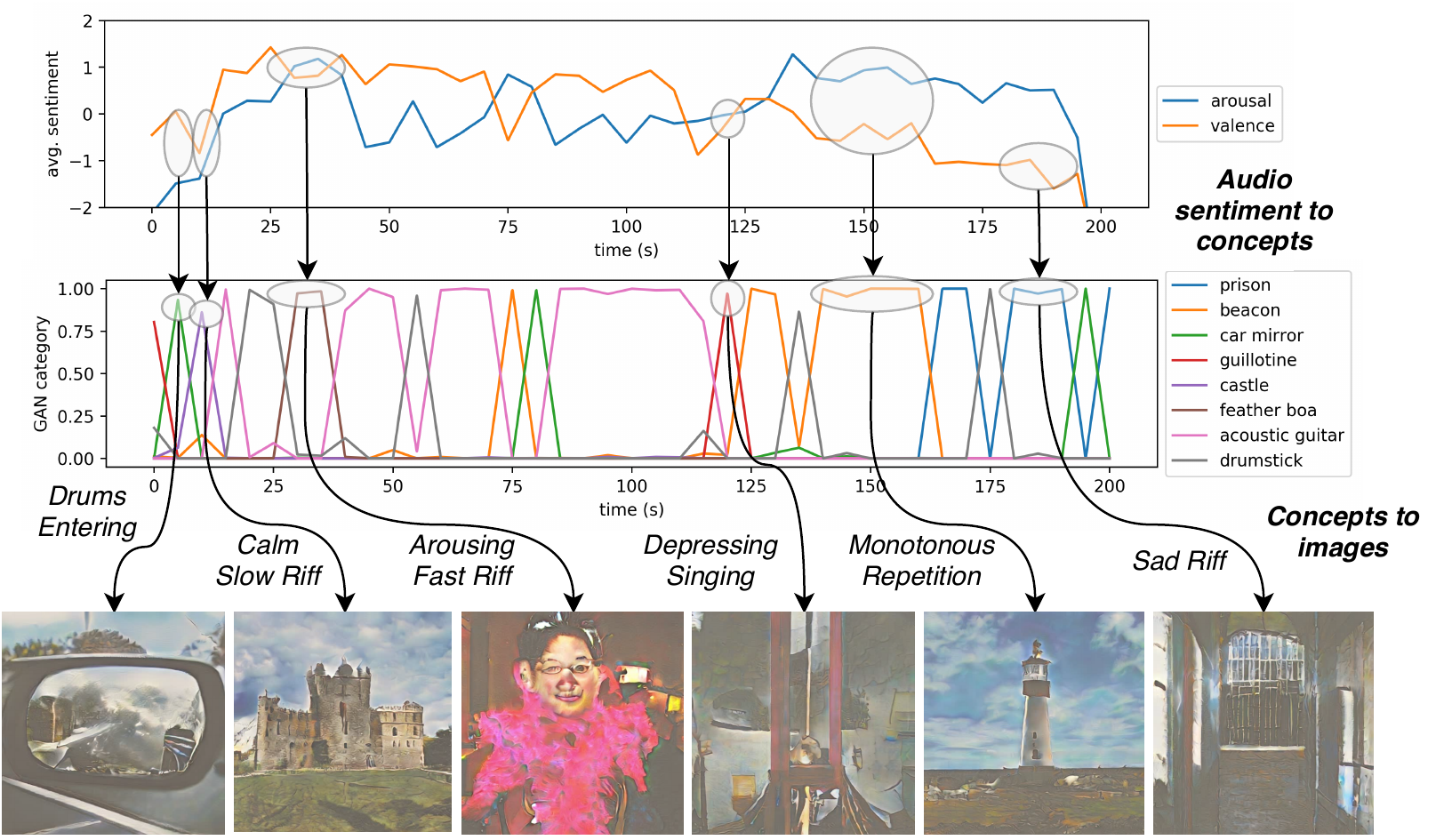}
	\end{center}

\scriptsize{Key frames selected along with annotations regarding the corresponding affective content of the song. For example, note the generated ``feather boa'' during the most arousing riff of the song and the transition to a ``prison'' as the valence of the song decreases.  Sample frames generated using the song ``Chop Suey!'' by ``System Of A Down''.}

\caption{Overview of the proposed method and a visualization example: music is translated into a visual story that express the same sentiment }

\label{fig:example}
\end{figure}

Our ability to perceive music is often reinforced by visual stimuli. For example, even in ancient Greek tragedies, music and dance performances were combined in the so-called \textit{stasima}, which were interludes, often emotionally-charged, between the main episodes, explaining and/or commenting on the episodes~\cite{taplin2003greek}. The advent of digital technology provided further opportunities toward integrating audio and visual content in novel ways. Most songs are now accompanied by videos that further reinforce their sentiment, while many music players are capable of generating visual patterns that are synchronized with each song, e.g., based on concurrent tones~\cite{ciuha2010visualization}, harmonic structures and other acoustic features~\cite{malandrino2015color,uehara2015pop}, or by detecting the sentiment of music~\cite{chen2008emotion,grekow2011emotion}.

The success of deep learning (DL) in various content generation and stylization tasks provides a powerful tool for tackling the aforementioned tasks.  For example, Generative Adversarial Networks (GANs)~\cite{goodfellow2014generative,karras2017progressive,brock2018large} are capable of synthesizing highly realistic visual content that has not been encountered during the training process, neural style transfer methods can re-paint images to match the style of reference images~\cite{luan2017deep}, or even to follow the style of well-known artists~\cite{gatys2015neural,wang2017multimodal}, while deep dreaming methods have demonstrated that neural networks can exhibit a behavior known as \textit{pareidolia} in humans, i.e., recognize and synthesize patterns on seemingly random data~\cite{mordvintsev2015inceptionism}. Despite the advanced capabilities provided by these approaches there has been no attempt,  to the best of our knowledge, to employ DL methods for generating emotionally-rich visual content that can match the sentiment expressed in various music compositions.

In this paper we provide the first, to the best of our knowledge, DL attribute-based music-to-image translation method, called \textit{deepsing}, that is capable of generating novel sentiment-aware visual content to accompany songs. That is, the synthesized visual stories are expected to induce  the same sentiment to the viewers as the music does, aiding in this way the primary purpose of music, i.e., communicating feelings. It is worth noting that even though we used sentiment as the basis of deepsing, the proposed method can be directly used to transfer any attribute across two different domains, providing a generic cross-modal translation methodology. The proposed method works  as follows: first the sentiment is extracted from a music track. Then, the sentiment is appropriately \textit{translated} and mapped into a space capable of generating visual content. Finally, a generator model, e.g., a GAN, is employed to generate the final content. This process, along with a few illustrative examples are provided in Fig.~\ref{fig:example}. It is worth noting that significant challenges are faced during this process, as we will further discuss through this paper, with the most significant one being the existence of multiple, equally viable, mappings between the sentiment space and the generator space. This often leads to unstable mappings, prohibiting us from training models for performing cross-modal translation. In this paper, we provide a novel way to translate the audio sentiment into images with similar sentiment, effectively overcoming this limitation. Compared with existing literature on music visualization~\cite{malandrino2015color,uehara2015pop,chen2008emotion,grekow2011emotion}, the proposed method employs a DL music-to-image translation approach for generating visual content that matches the sentiment of corresponding music segments. Note that, to the best of our knowledge, none of the existing methods is capable of generating meaningful visual content that can match the sentiment of songs, with most of them being limited to either simplistic animations, often with little-to-zero artistic value, e.g., \cite{malandrino2015color,uehara2015pop},  or using collections of photos to generate slide-shows~\cite{chen2008emotion}. The proposed method does not only provide a novel way to generate sentiment-enriched visual stories, but opens a whole new research area for music-to-image translation with several high-value applications in many different domains. To aid research on this domain, we provide an open-source implementation of the method proposed in this paper at \url{http://soon.to.be.available}.

The rest of the paper is structured as follows. First, the problem of music-to-image translation is formally defined and the proposed method is derived in Section~\ref{sec:proposed}. Then, the experimental evaluation of the proposed method is provided in Section~\ref{sec:experiments}. Finally, conclusions are drawn and future research directions are discussed in Section~\ref{sec:conclusions}.
 
\section{Proposed Method}
\label{sec:proposed}

In this Section we formally define the problem of music-to-image translation, provide the proposed pipeline for this process and demonstrate how this approach can be used for generating sentiment-aware visual stories. Then, we analytically derive the employed neural attribute translation method, which lies at the heart of the proposed pipeline. Finally, we introduce an attribute-based neural stylization approach, that can further improve the relevance of the generated images with the given attributes, while also allowing the users to adapt this process to their preferences.

\subsection{Music-to-Image Translation}
Let $\mathbf{x} \in \mathbb{R}^{N_s}$ be an audio segment with $N_s$ samples. Also, let $f_a(\mathbf{x}) \in \mathbb{R}^{N_a}$ be an \textit{audio attribute estimator}, where $N_a$ is the dimensionality of the attribute space. The attribute space describes a specific property of the music that we want to maintain in the visual domain. For example,  $f_a(\cdot)$ can extract information regarding the sentiment of the music, e.g., its valence and arousal, or information regarding the semantics of the lyrics. We also introduce an attribute extractor for the visual domain $g(\mathbf{y}) \in \mathbb{R}^{N_a}$, where $\mathbf{y} \in \mathbb{R}^{W\times H \times C}$ is a $C$-channel image of dimensions $W\times H$. The\textit{ visual attribute estimator} extracts the same attributes as the audio attribute estimator, but operates on images instead of audio. For example, it can extract information regarding the sentiment that  an image induces to its viewers, or attributes regarding the semantic content of images, e.g., categories of the objects that appear in an image. These two attributes estimators should \textit{aligned}, i.e., extract the same attributes and operate on the same output space.  We also define a divergence metric $\mathcal{D}(\mathbf{t}_1, \mathbf{t}_2)$ for measuring the dissimilarity between two attribute vectors $\mathbf{t}_1, \mathbf{t}_2 \in \mathbb{R}^{N_a}$. For example, $\mathcal{D}(\cdot)$ can be defined using $l^2$ norm, i.e., $\mathcal{D}(\mathbf{t}_1, \mathbf{t}_2) = \lVert \mathbf{t}_1 - \mathbf{t}_2 \rVert_2$.

The problem of music-to-image translation can be then defined as:
\begin{mydef} [Music-to-Image Translation]
Given an audio segment $\mathbf{x}$, generate an appropriate image $\mathbf{y}$ so as:
\begin{equation}
\label{eq:itm}
\mathbf{y} = \arg_\mathbf{y} \min \mathcal{D}\left(f_a(\mathbf{x}), g(\mathbf{y})\right),
\end{equation}
for an appropriately defined divergence metric $\mathcal{D}(\cdot)$.
\end{mydef}
Therefore, music-to-image translation aims to generate images that will carry the same attributes as the corresponding audio segments do, given the estimators $f_a(\cdot)$ and $g(\cdot)$. These estimators can be trivially fitted using annotated datasets for the corresponding domains. As already discussed, in this paper we focus on {sentiment-oriented} music-to-image translation, aiming to generate images that will induce the same sentiment to a viewer as the sentiment of the music to be translated. Therefore, the audio attribute estimator extracts the sentiment of the music, while the visual attribute estimator extracts the sentiment of the corresponding images. Please refer to Section~\ref{sec:experiments} for more details on the datasets and models used for training these estimators.  It is also worth noting that there is a ``1-to-N'' correspondence between each sentiment and the possible visual outputs that can induce this sentiment, i.e., there are many different visual stimuli that can lead to the same sentiment. Therefore, it is often critical to restrict the search space according to preferences of users, as we also discuss later in this Section, in order to produce meaningful and consistent results.

\begin{figure}[t]
	\begin{center}
		\includegraphics[width=0.99\linewidth]{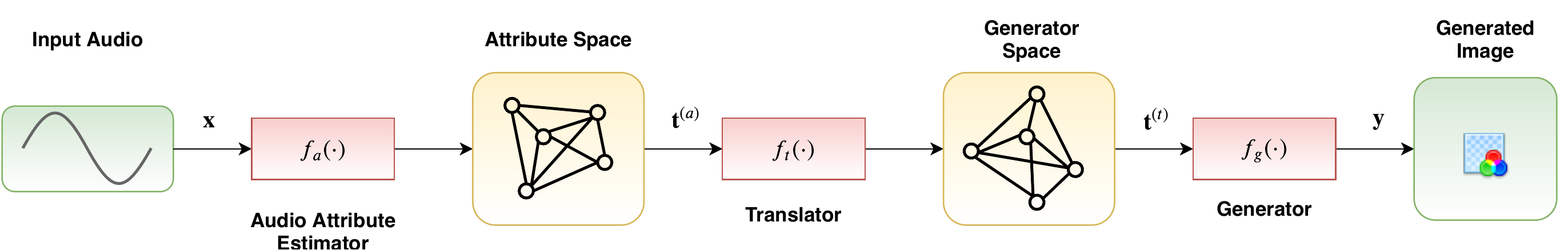}
	\end{center}
	\caption{Proposed audio-to-image translation pipeline}
	\label{fig:pipeline-proposed}
\end{figure}

The proposed pipeline for music-to-image translation is shown in Fig.~\ref{fig:pipeline-proposed}. First, the audio attribute extractor is employed to extract the attribute vector $\mathbf{t}^{(a)} = f_a(\mathbf{x})$ for a given audio segment $\mathbf{x}$.  The proposed method aims to \textit{translate} this vector into an appropriate \textit{generator vector} $\mathbf{t}^{(t)}$, that can be fed into a generator model $\mathbf{y} = f_g(\mathbf{t}^{(t)})$, in order to acquire an image with the same attributes as the corresponding audio segment. In this work, we employ a GAN for generating the final images from the intermediate generator vectors. The gist of the proposed method is to learn an appropriate translation model  $\mathbf{t}^{(t)} = f_t(\mathbf{t}^{(a)})$ that will map the attribute vectors to the appropriate generator vectors, ensuring that the divergence between the audio attributes and the visual attributes of the generated image will be minimized, as required by (\ref{eq:itm}). In other words, the translation model must learn how to appropriately ``control'' the generator in order to generate images with the  attributes provided by the audio attribute estimator.

\subsection{Neural Attribute Translation}

In this work, we propose to learn how to translate the attribute vectors to generator vectors by learning how to inverse the visual attribute estimator $g(\cdot)$. Therefore, the optimization problem given in~(\ref{eq:itm}) can be reduced to:
\begin{equation}
\mathbf{y} = f_g\left(\mathbf{t}^{(t)}\right), \text{ where }  \mathbf{t}^{(t)} = f_t\left(f_a(\mathbf{x})\right), 
\end{equation}
and 
\begin{equation}
\label{eq:new-optim}
f_t = \arg_{f_t} \min \mathcal{D}\left(f_a(\mathbf{x}), g\left(f_g\left(\mathbf{t}^{(t)}\right)\right)\right).
\end{equation}
That is, the image $\mathbf{y}$ can be trivially generated after fitting an appropriate translation model $f_t(\cdot)$ that minimizes the attribute divergence between audio attributes and visual attributes. To efficiently learn the translation model $f_t(\cdot)$ we propose sampling the generator space in order to collect training data in the form of pairs $(\mathbf{t}^{(t)}_i, \mathbf{\tilde{t}}_i^{(a)})$,
where $\mathbf{\tilde{t}}^{(a)}$ is the estimated attribute vector of the generated image, as calculated using the visual attribute estimator:
\begin{equation}
\label{eq:id}
\mathbf{\tilde{t}}_i^{(a)} = g(f_g(\mathbf{t}^{(t)}_i)).
\end{equation} It is worth noting that for the case of GANs, sampling the generator space is easy, since GANs are typical trained to generate images from a Gaussian distribution~\cite{brock2018large}. Then, the translation model can be trivially learned to minimize the following loss:
\begin{equation}
\label{eq:loss}
\mathcal{L} = \sum_{i=1}^N \lVert f_t(\tilde{\mathbf{t}}_i^{(a)}) - \mathbf{t}^{(t)}_i \rVert_2,
\end{equation}
for $N$ pairs of generator-attribute vectors. That is, the translation model learns how to map the attribute vectors, as induced by the visual modality, to the generator vectors that should be used to generate the images with the corresponding attributes. Gradient descent can be used to fit the translation model, i.e., $\Delta \mathbf{W} = -\eta \frac{\partial \mathcal{L}}{\partial \mathbf{W}}$, 
where $\eta$ is the learning rate and the notation $\mathbf{W}$ is used to refer to the parameters of the translation model, which is typically a deep neural network. It is trivial to verify that learning the translation model in this way ensures that (\ref{eq:new-optim}) is indeed minimized, since
\begin{equation}
g\left( f_g\left( f_t(\mathbf{t}^{(a)}) \right) \right) =  \mathbf{t}^{(a)},
\end{equation}
by (\ref{eq:id}) and assuming that $f_t = \arg_t \min \mathcal{L}$. Therefore, the divergence given in (\ref{eq:new-optim}) reduces to $\mathcal{D}\left(\mathbf{t}^{(a)},g\left( f_g\left( f_t(\mathbf{x}^{(a)}) \right) \right)\right) = \mathcal{D}\left(\mathbf{t}^{(a)}, \mathbf{t}^{(a)}\right) = 0$.

\begin{figure}[t]
	\subfloat[\label{fig:translator-training}]{
		\includegraphics[trim={0cm 1.9cm 0cm 0cm }, clip,width=0.3\linewidth]{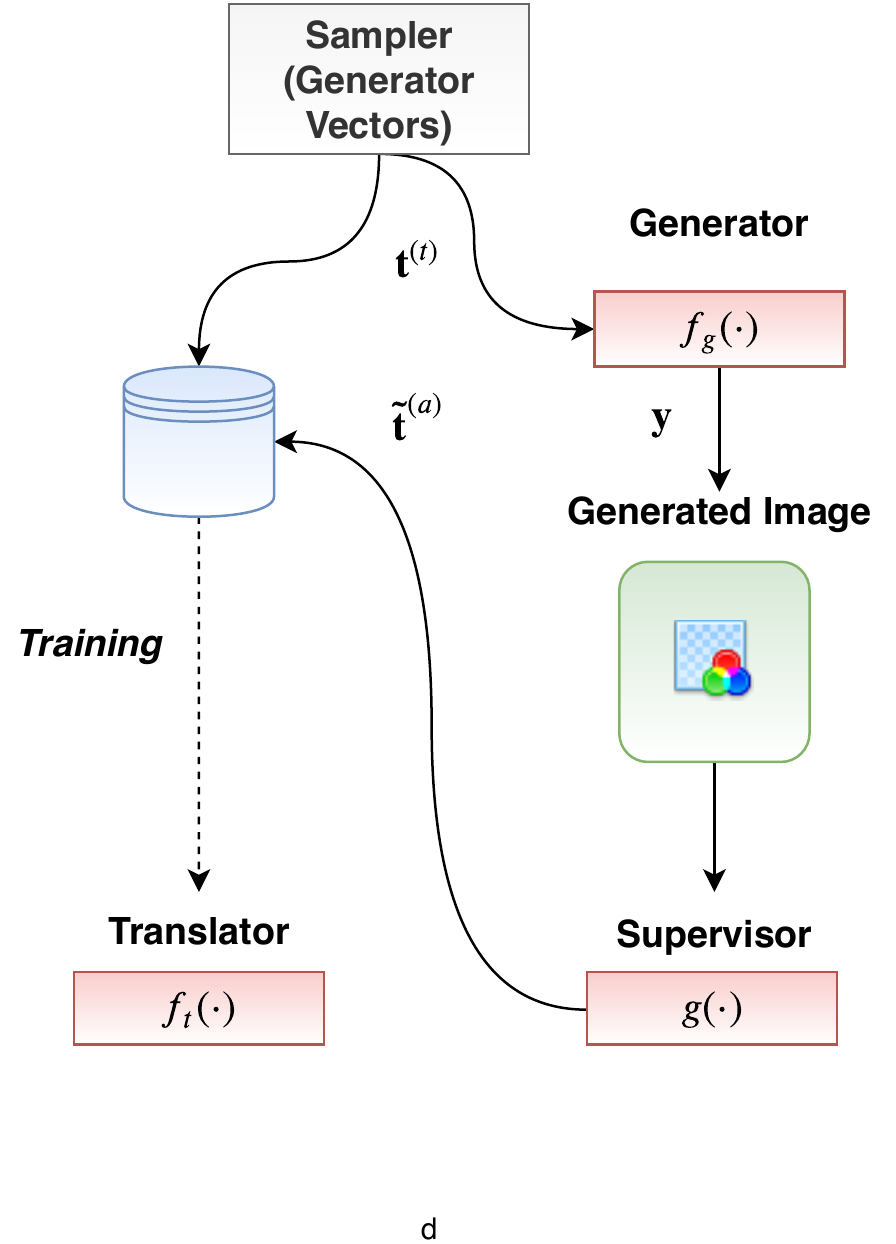}
	}
	\subfloat[\label{fig:gan_issue}]{
		\includegraphics[trim={0.cm 0cm 0.cm 0.5cm }, clip,width=0.3\linewidth]{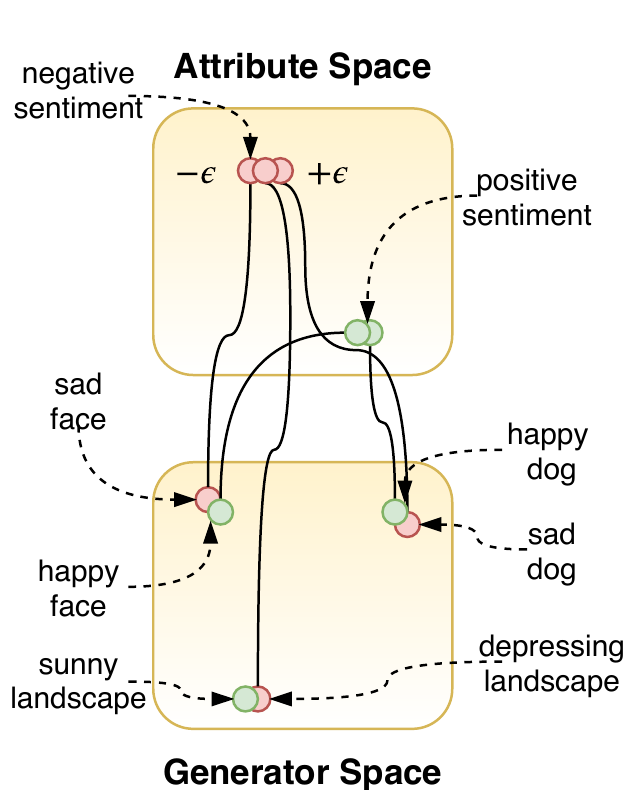}
	}
	\subfloat[\label{fig:gan-example}]{
		\raisebox{0.17\height}{
			\includegraphics[trim={1.9cm 1.2cm 0.5cm 0.8cm }, clip, width=0.38\linewidth]{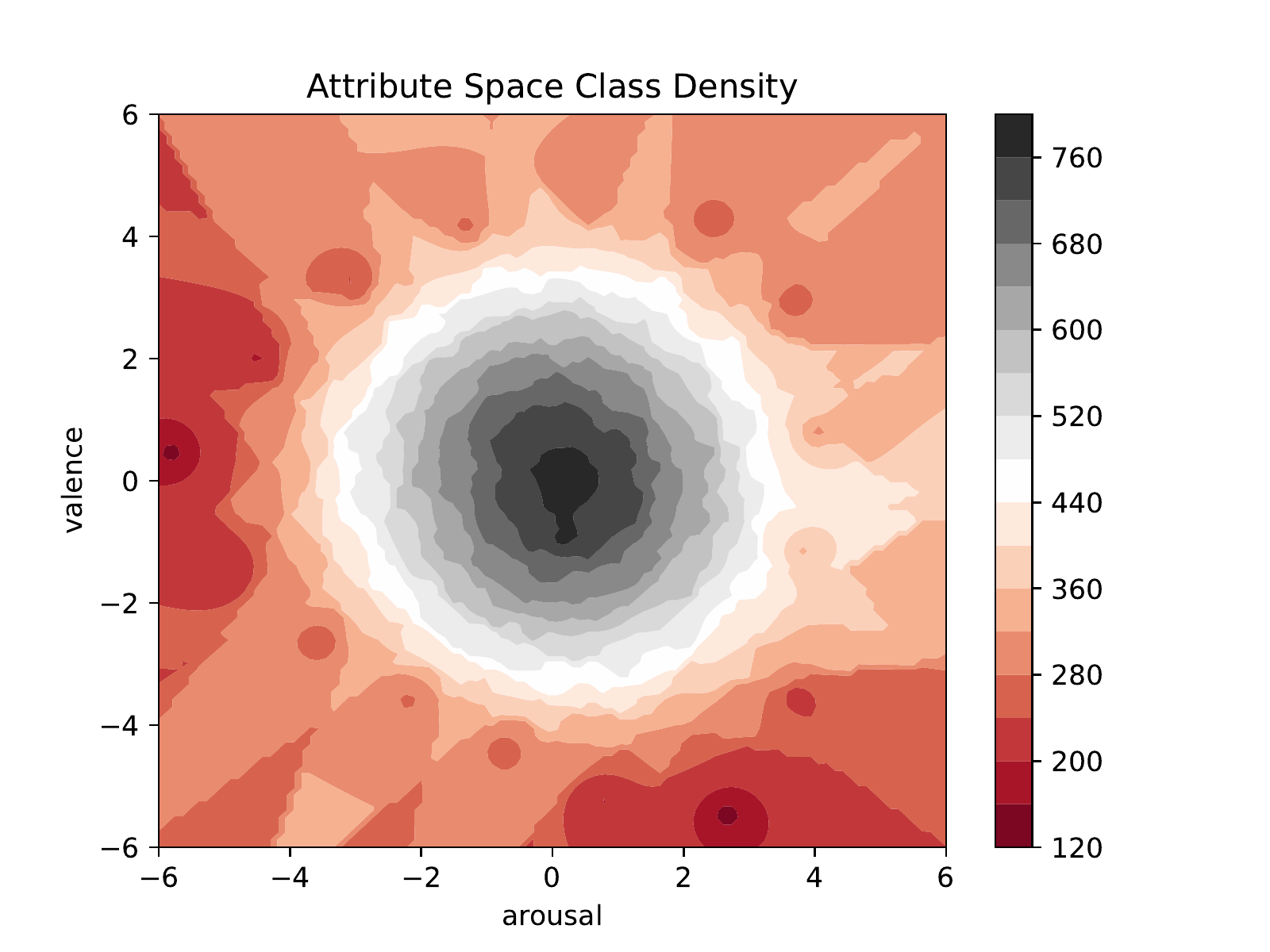} 	    
		}
		
	}	
	\caption{Proposed method for fitting the translation model (Fig. 2a), an unstable  mapping can occur between the attribute and generator spaces (Fig. 2b), and a toy example demonstrating the unstable mapping for a GAN with 1000 classes (Fig. 2c). }
	\label{fig:pipeline}
\end{figure}

The proposed method for learning the translation model is summarized in Fig.~\ref{fig:translator-training}. A sampler is employed to generate multiple generator vectors by drawing from an appropriate distribution, typically a Gaussian one. Then, these vectors are used to generate images, from which we extract visual attributes with the visual attribute estimator. The visual attribute estimator is also called \textit{supervisor} in Fig.~\ref{fig:translator-training}, since it supervises the training of the translation model. After gathering enough pairs of sampled generator vectors and the corresponding attribute vectors, the translator is fitted. 

The aforementioned process theoretically guarantees that the optimal translator will be obtained, if enough samples are acquired and an appropriate translation model, with  respect to its learning capacity, is used. However, in practice we observed that it was very difficult to fit such translation models, especially if GANs with very complex generator spaces are used, e.g., GANs capable generating 1000 classes~\cite{brock2018large}. To understand why this happens, we have to consider the mapping between the (sub)-classes produced through the GAN and the attribute space. For example, consider the case of an attribute space that describes the sentiment of an image and a generator space from which multiple classes will be generated, as shown in~ Fig.~\ref{fig:gan_issue}. Note that for very small variations in the sentiment (denoted by $\epsilon$ in Fig.~\ref{fig:gan_issue}) very large changes can occur in the mapping with the generator space, i.e., small variations of the negative sentiment can lead to many significantly different, yet equally negative, sub-classes. It is worth noting that this is the typical behavior of a {chaotic system}~\cite{tsuda2001toward}, explaining the difficulties in fitting such translation models. At the same time, also note that large variations in the sentiment might lead to very slight variations in the generator space, e.g., happy/sad dogs are closer compared to happy dogs and happy faces.  This behavior was also experimentally confirmed for the GAN used in the experiments conducted in this paper, as shown in Fig.~\ref{fig:gan-example}. The plot shown in Fig.~\ref{fig:gan-example} was generated by clustering the sound attribute space and then measuring the number of GAN classes mapped in each cluster. Note that a very large number of classes (often more than the half, i.e., more than 500) are mapped in most of the clusters, rending the mapping especially unstable.

\begin{algorithm}[t]
	\caption{Calculating Stable Attribute Views }\label{algo:attribute_view}
	\begin{algorithmic}[1]
		\Procedure{AttributeView}{$N_K$, $N_S$}
		\State Sample the generator space and generate $N$ pairs $(\mathbf{t}^{(t)}_i, \mathbf{\tilde{t}}_i^{(a)})$
		\State Cluster attribute pairs  into $N_K$ clusters according to $\mathbf{\tilde{t}}_i^{(a)}$
		\State Start with any empty training set $\mathcal{X}_{train} = []$
		\For {each cluster} {}
		\parState{Sample one GAN category $c$ from each cluster according to the probability of observing each class in the generator space }
		\State {Add every instance $(\mathbf{t}^{(t)}_i, \mathbf{\tilde{t}}_i^{(a)})$ of category $c$ to $\mathcal{X}_{train}$ }
		\EndFor
		
		\For {each category $c$ in $\mathcal{X}_{train}$} {}
		
		\State Cluster the attribute vectors $\mathbf{\tilde{t}}_i^{(a)}$ of each category  $c$  into $N_S$ sub-categories
		\State Let $\mathcal{X}_i$ contain every pair $(\mathbf{t}^{(t)}_i, \mathbf{\tilde{t}}_i^{(a)})$ that belong to the $i$-th sub-category
		\For {each sub-category $i$} {}
		\State Calculate a smooth attribute vector $\mathbf{t}^{(a)}_{target} = \frac{1}{|\mathcal{X}_i|} \sum_{(\mathbf{t}^{(t)}, \mathbf{t}^{(a)}) \in \mathcal{X}_i}  \mathbf{t}^{(a)}$
		\State Calculate a smooth generator vector  $\mathbf{t}^{(t)}_{target} = \frac{1}{|\mathcal{X}_i|} \sum_{(\mathbf{t}^{(t)}, \mathbf{t}^{(a)}) \in \mathcal{X}_i}  \mathbf{t}^{(t)}$
		\parState {Replace each  $(\mathbf{t}^{(t)}_i, \mathbf{\tilde{t}}_i^{(a)})$ pair in $\mathcal{X}_i$ by its corresponding smoothed target $(\mathbf{t}^{(t)}_{target}, \mathbf{t}^{(a)}_{target})$}
		\EndFor
		\EndFor

		\EndProcedure
	\end{algorithmic}
\end{algorithm}

To overcome this limitation, we propose creating a stable \textit{attribute view} of the mapping between these two spaces. Therefore, instead of trying to match each attribute with every possible generator vector that can induce this attribute, we propose using only a part of the generator space. This limits the number of images that can be generated by $f_g(\cdot)$, effectively providing a specific view on the given attributes. Multiple views that can  lead to the same attributes can be acquired by small perturbations of the attribute space. 

The algorithm employed for acquiring a view of the generator space is provided in Algorithm~\ref{algo:attribute_view}. In this work, we assume that a class-based GAN is employed. However, the proposed method can be also extended to handle any kind of GANs. First, the generator space is sampled and the visual attribute vectors are extracted (line 2). Then, these vectors are clustered into $N_K$ categories (line 3). For each cluster, we sample a GAN category with probability proportional to its cardinality in the cluster. Then, the latent vectors for all other categories in this cluster are discarded (lines 4-7). This process allows to effectively keep only one GAN category per cluster, leading to a smoother and much stabler matching, since every remaining attribute vector in each cluster is mapped to the same class. Note that, as demonstrated in Fig.~\ref{fig:gan-example}, for a GAN capable of generating images belonging to 1,000 different categories, each initial cluster could be mapped to more than 500 different categories.  Therefore it is expected that this process will  improve the stability of the mapping, as we indeed experimentally demonstrate in Section~\ref{sec:experiments}.

After selecting the categories to be used, then we further cluster each category to detect sub-cluster that express different attributes (lines 8-14), e.g., different sentiments are detected in the case of our application. To this end, we cluster the attribute vectors for each category and then we calculate the centroid of each sub-cluster and each generator vector (lines 12-14), further smoothing the mapping. This process effectively allows to keep only the most prominent mappings between attribute and generator vectors.  This process can seemingly reduce the variation during the content generation process. However, this is not expected to be a significant issue since a) a large number of sub-clusters is typically used (i.e., $N_S > 10$), while b) during the content generation and after calculating the vector $\mathbf{t}^{(t)}$ for a given class, a small (and easily controllable compared to the translation mapping) amount of noise can be optionally used to increase the variation of the generated content, if needed. Finally, note that users can also supply a number of categories that wish to use during the content generation. In this case, these categories can be directly used, skipping the first clustering step (lines 3-7) in Algorithm~\ref{algo:attribute_view}.

\subsection{Hyper-stylization}

Even though GANs can offer a satisfactory degree of variation for the generated content, they currently mostly fail to also simultaneously stylize the generated images according to the requirements of the users. Therefore, to further improve the style variation of the generated content, we propose using a few user-supplied styles to stylize the generate images.  This process allows the users to adapt the generated visual stories to their preferences. In this work we opt for employing a universal style transfer approach, e.g.,~\cite{li2017universal}, allowing for using just one image per style. Therefore, after the user supplies the style images, these styles can be directly mapped to the attribute space using the visual attribute estimator $g(\cdot)$. Then, the style image that is closer to the current attributes can be employed to stylize the generated content. Examples of this process are also provided in Section~\ref{sec:experiments}. Finally, note that for human-friendly attributes, such as sentiment-based ones, e.g., arousal, valence, etc., the users can also manually set the thresholds that should be used for this process. 

\section{Experimental Evaluation}
\label{sec:experiments}

\subsection{Experimental Setup}

In this work, we focused on generating sentiment-oriented visual stories from music. To this end, we trained the both the audio and visual attribute estimators to regress the valence and arousal of audio and visual stimuli respectively. It is worth noting that this choice was largely dictated by the current availability of open datasets for training sentiment-related attribute extractors. Any kind of attributes/features, can be also used to this end, e.g., sentiment embedding can be used instead of the employed valence-arousal features~\cite{tang2015sentiment}.

For developing the  audio attribute estimator we extracted a) 40 mel frequency cepstral coefficients (MFCC)~\cite{logan2000mel}, b) chroma energy normalized (CENS) features~\cite{muller2011chroma}, and c)  tempogram features~\cite{grosche2010cyclic}. For all these features, we used the default parameter/extraction setup provided by the librosa library~\cite{brian_mcfee_2019_3478579}. One feature vector containing the concatenation of these three different features was extracted every 500ms. The extracted features were then fed into a neural regressor consisting of one hidden layer with 256 neurons (using sigmoid activations, which led to consistently better regression performance) and one output regression layer with 2 neurons. The regressor was trained to predict the valence and arousal of music segments of 500ms using the Database for Emotional Analysis of Music (DEAM)~\cite{alajanki2016benchmarking}. For training the visual attribute extractor we employed a MobileNetV2 model~\cite{sandler2018mobilenetv2}, trained to regress the valence and arousal using the Open Affective Standardized Image Set (OASIS)~\cite{kurdi2017introducing}. The sound attribute estimator was trained for 30+20 training epochs using a learning rate of $10^{-4}/10^{-5}$. The visual attribute estimator was pretrained on the Imagenet dataset and then fine-tuned (after replacing the last regression layer) for 50+10 epochs using the same learning rates ($10^{-4}/10^{-5}$). The Adam method was used for optimizing the models~\cite{kingma2014adam}. The visual attribute space and the audio attribute space were aligned by performing z-score normalization. For the visual space we employed the statistics of the corresponding training dataset, while for the audio space we use song-level z-score normalization, except from the third song used for the conducted experiments, where using the statistics of the whole dataset led to a wider variety of generated concepts.

A pretrained BigGAN~\cite{brock2018large} model was used for generating images of $512 \times 512$ pixels. The translation model consists of two hidden hidden layers with 64 and 256 neurons respectively which branch out into two streams: a) a 1000 classification (softmax) layer used for predicting the category that will be used by the GAN, and b) a 128 fully connected layer with no activation function used for predicting the latent vector to be fed to the GAN. The translation model was trained for 200 epochs with a learning rate of 0.001. Gaussian noise with $\sigma=0.1$ was added to the translated latent vectors, to increase the variation of the generated content. Finally, for performing stylizations we employed the Whitening Coloring Transform (WCT)-based stylization method~\cite{li2017universal}. The first four encoder-decoders were used for the stylization process, while the stylization blending factor was set to 0.1.

For the three song used for studying the behavior of the proposed method, the concept classes were manually selected, and the algorithm presented in Section~\ref{sec:proposed} was used to automatically match the sentiment of each song to the available concepts. It is worth noting that no human intervention was allowed during the process of generating the visual story. Therefore, the matching between the available concepts and classes was performed fully autonomously.

\subsection{Experimental Results}

First, we examine three visual stories generated by the proposed method for three well known songs. The first one was provided in Fig.~\ref{fig:example}, where the song ``Chop Suey!'' was used. In the upper plot we provide the average arousal and valence over the music intervals (5 seconds each) used for the content generation, while in the subsequent plot we provide the activations of the translator model. We examine the behavior of the proposed method at six selected points of interest. 
For the first two points of interest the arousal is low, while the valence is near to zero. Indeed, these points correspond to a more ``relaxed'' part of the song, where the drums are just entering and a calm slow riff accompanies the song. The generated images match this sentiment, since they are quite neutral and with low arousal (a car mirror and castle in the country side). On the other hand, the third point of interest corresponds to one of the most arousing parts of the song, where a very fast and exciting riff is played. The proposed method responds to this by generating a bright pink feature boa. Then, the arousal is reduced and a few matching classes are used for the visualization (e.g., acoustic guitar). Then, a guillotine is visualized at a critical transition point of the song, at the fourth point of interest), where both the singing style and the lyrics are less positive (the lyrics actually refer to suicide at this point). Then, the valence is slowly decreasing by a monotonous repetition of the same musical phrase, at the fifth point of interest, which is accompanied by a visualization of an isolated beacon, creating the sense of loneliness. The drop in the arousal, which is especially evident at this point in the song, concludes by generating frames containing  a prison, at the final point of interest.

\begin{figure}
	\begin{center}
		\includegraphics[trim={0cm 0cm 0cm 0cm},clip, width=0.99\linewidth]{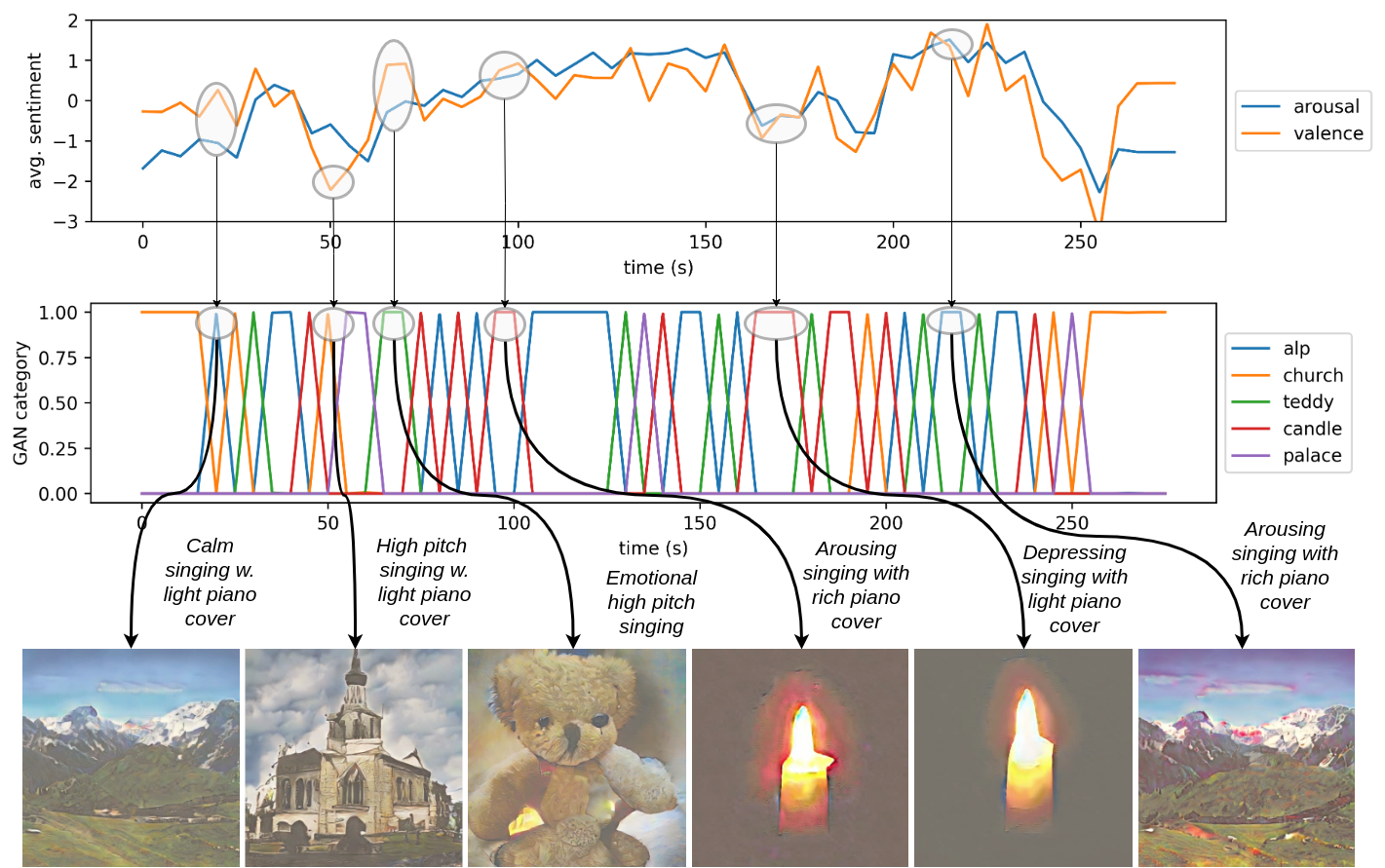}

		\footnotesize{ Sample frames generated using the song ``Take Me To Church (Cover)'' by ``Postmodern Jukebox''.}
	\end{center}
	\vspace{0.1em}
	\begin{center}
		\includegraphics[trim={0cm 0cm 0cm 0cm},clip, width=0.99\linewidth]{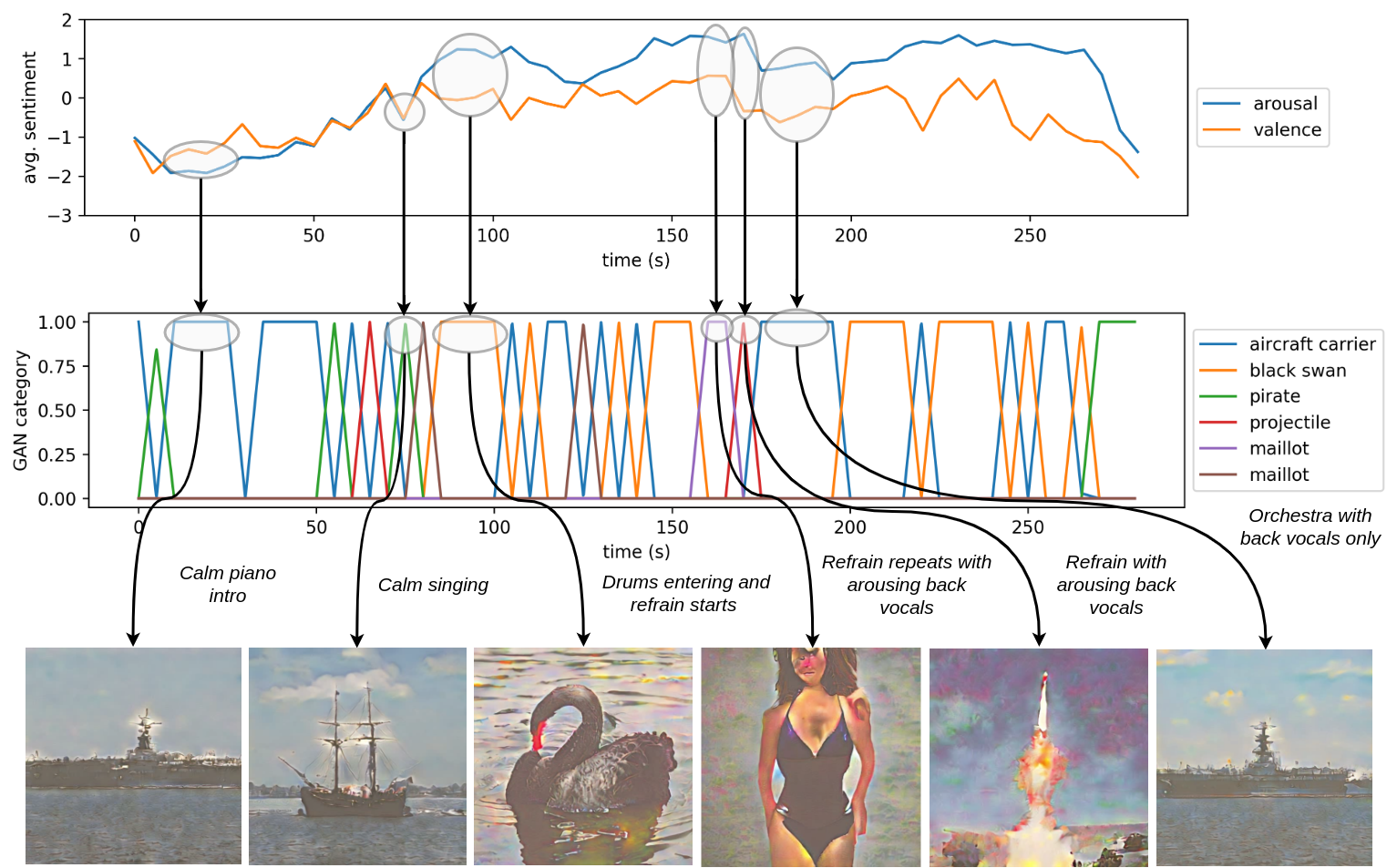}
		
		\footnotesize{ Sample frames generated using the song ``Skyfall'' by ``Adele''.}
	\end{center}	
	
	\caption{Two visual stories generated for two different songs }
	
	\label{fig:more:examples}
\end{figure}

The next song used for evaluating the proposed method is the ``Take Me To Church'' originally performed by ``Hozier''. As before, we visualize some points of interests in the generated visual story in Fig.~\ref{fig:more:examples}. The song begins with a calm and slow singing style, accompanied with a light piano cover. At this point a low arousal, neutral valence scenery of a mountain is generated. After a few seconds the valence reduces, with the singer increasing the pitch, leading to the generation of a church. The negative valence of the generate image is further reinforced by the cloudy sky. Then, at the third point of interest, the singer still maintains a high pitch, but with a significantly warmer tone. At this point a more positive teddy bear is generated. Then, as the intensity of the song rises, a warmly colored candle is generated. It is worth noting that a similar candle is also used to visualize another point of interest, which has significantly lower valence and arousal. Indeed, this is illustrated in the generated images, since the candle of the fourth point of interest has an overall more positive appearance compared to the one used in the firth point of interest. This behavior confirms the ability of the proposed method to generate images that can highlight the subtle emotional differences between different instances of the same class. This is also confirmed in the last generated instance, where a more positive scenery was generated compared to the one used in the first frame.

The third visual story, depicted in the lower part of Fig.~\ref{fig:more:examples}, was generated using the song ``Skyfall'' by ``Adele''. Dominant element in this visual story is sea, which exists in two of the most frequently generated classes, i.e., ``aircraft carrier'' and ``pirate ship''. Frames containing these classes are generated mostly during the lower intensity parts of the song, especially in the beginning, where both the valence and the arousal is low. During the arousal build up phase, where the drums enter and the more intense refrain begins, a black swan is generated, while the arousal peaks during the fourth and fifth frame, with more arousing content being generated.

\begin{figure}[t]
	\begin{center}
		\includegraphics[trim={0cm 0cm 0cm 0cm},clip, width=0.99\linewidth]{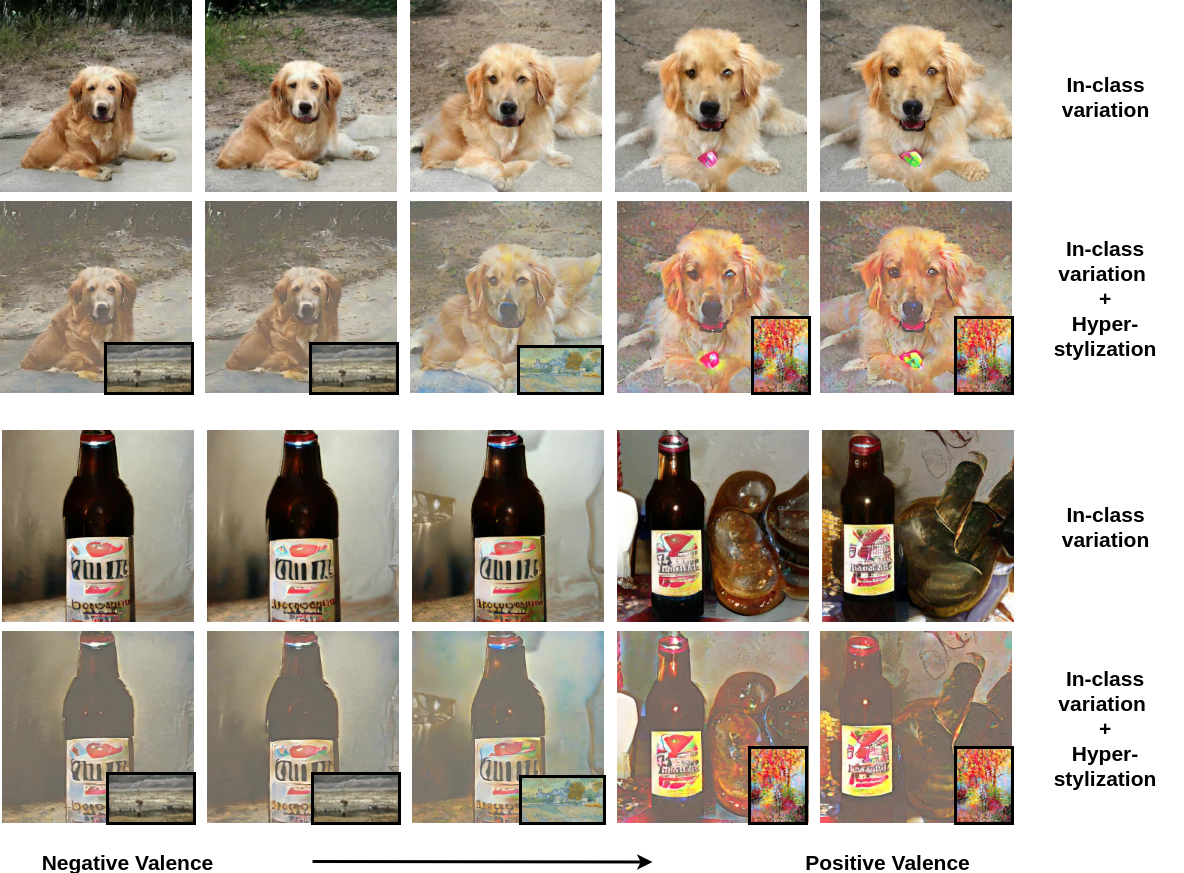}

	\end{center}

	\caption{Examining the ability of the translator model to discover sentiment-rich regions of the generator space, along with the effect of the proposed hyper-stylization approach.}
	
	\label{fig:ablation}
\end{figure}

Finally, the effect of the proposed hyper-stylization method is demonstrated in Fig.~\ref{fig:ablation}, where images of one class, ranging from the ones with the most negative valence to the ones with the most positive one, were generated. Two classes were used for this experiment: ``dog'' and ``beer''. First, note that even though the employed GAN was not trained to generate images with different valance, the proposed translation approach manages to reveal some meaningful valance characteristics. For example, dogs with more positive valence tend to have their mouth open, while beer bottles that co-exist with other items were also classified as more positive. The proposed hyper-stylization approach can further improve the valence of the generated images by employing three user-defined style images (depicted in the lower right part of the images). It is evident that the proposed hyper-stylization process can indeed significantly improve the matching between the desired valence and the valence of the generated images.  It is worth noting that the same style images were used for all the conducted experiments in this paper, while the threshold for neutral sentiment was set to 0.5: an image with an average sentiment (valence and arousal) lower than -0.5 was considered as negative, while an average sentiment higher than 0.5 was considered positive.

\section{Conclusions}
\label{sec:conclusions}

In this paper we introduced \textit{deepsing}, a deep learning method for generating sentiment-aware visual stories by performing cross-modal translation from the audio domain. The proposed method works by first extracting the sentiment of a music track, which is then appropriately translated into a space, from which a GAN can be employed for generating the frames of the visual story. This process was proven to be especially challenging, since the mapping between the sentiment space and the generator space is unstable, requiring special care to ensure the stability of the matching. To further enhance the quality of the visual stories we employed a hyper-stylization approach, that performs attribute-aware stylization. The ability of the proposed method to produce meaningful visual stories was demonstrated using three well-known modern songs.  

The results obtained in many cases were quite spectacular, matching the generated visual content to the audio one in a very satisfactory way.  Despite this, there are several open research question and challenges associated with the proposed method. First, defining a meaningful evaluation metric for  computer generated content is among the greatest challenges in measuring the performance of such approaches. Furthermore, for each cluster formed in the attribute space there are many potential candidate classes. Selecting the class to use for the content generation according to a semantic similarity measure with the rest of the selected classes, instead of performing cardinality-based sampling, is expected to improve the semantic consistency of the generated story. Finally, aligning the attribute spaces of the audio attribute estimator and visual attribute estimator with more robust approaches, that can also ensure the diversity of the generated content, is expected to further improve the quality of the visual stories.

\bibliographystyle{splncs}
\bibliography{egbib}
\end{document}